\documentclass{article}
\usepackage{spconf,amsmath,epsfig}

\usepackage{cite}

\usepackage{graphicx}
\graphicspath{{Figures/}}
\DeclareGraphicsExtensions{.pdf,.jpeg,.png}

\usepackage{amsmath}
\usepackage{algorithm}
\usepackage{algpseudocode}

\usepackage{gensymb}
\usepackage{subfloat}

\usepackage{array}

\usepackage{multirow}
\usepackage{amsfonts}
\usepackage{stfloats}

\usepackage{url}
\usepackage[export]{adjustbox}
\usepackage{textcomp}
\usepackage{cite}

\usepackage{caption}
\usepackage[font=footnotesize]{subcaption}
\usepackage{breqn}
\usepackage{lipsum}
\usepackage{framed,multirow}
\usepackage{booktabs}
\usepackage{url}
\usepackage{xcolor}

\usepackage{comment}

\usepackage{algorithm}
\usepackage{eqparbox}
\usepackage{soul} %

\usepackage{graphicx}
\usepackage{amsmath,amssymb} %
\usepackage{bm}
\usepackage{booktabs}
\usepackage{rotating,multirow}
\usepackage{wrapfig}

\usepackage{tikz}
\usepackage{pgfplots}
\pgfplotsset{width=7cm,compat=1.16} 
\usepackage[export]{adjustbox}
\usepackage[inkscapearea=page]{svg}
\setsvg{inkscape=inkscape -z -D,svgpath=fig/}

\algnewcommand{\LeftComment}[1]{\Statex \# #1}

\makeatletter
\newcommand{\dummylabel}[2]{\def\@currentlabel{#2}\label{#1}} %
\makeatother

\graphicspath{{./Images/}{./Figures/result/}{./Images/result/}}

\newcommand{\TODO}[1]{{\color{blue}\textbf{TODO: }#1}}

\usepackage{algpseudocode}

\begin{document}

    \title{\TODO{Many title candidates are possible, but only the last taken}}
    \title{Leveraging feature communication in federated learning \\ for remote sensing image classification}
    \name{Anh-Kiet Duong$^{*,\dagger}$, Hoàng-Ân Lê$^ *$, Minh-Tan Pham$^ *$~\thanks{This research work is supported by the SAD 2021 ROMMEO project (ID 21007759) and the ANR AI chair OTTOPIA project (ANR-20-CHIA-0030). The study program pursued by the first author is sponsored by the ``Vingroup Science and Technology Scholarship Program for Overseas Study for Master's and Doctoral Degrees''.}} %
    \address{$^*$IRISA, Université Bretagne Sud, UMR 6074, 56000 Vannes, France \\
    $^{\dagger}$ Université de Limoges, 87060 Limoges, France\\
    \tt\small anh-kiet.duong@etu.unilim.fr, \{hoang-an.le,minh-tan.pham\}@irisa.fr}
	\maketitle
	
\begin{abstract}
    
In the realm of Federated Learning (FL) applied to remote sensing image classification, this study introduces and assesses several innovative communication strategies. Our exploration includes feature-centric communication, pseudo-weight amalgamation, and a combined method utilizing both weights and features. Experiments conducted on two public scene classification datasets unveil the effectiveness of these strategies, showcasing accelerated convergence, heightened privacy, and reduced network information exchange. This research provides valuable insights into the implications of feature-centric communication in FL, offering potential applications tailored for remote sensing scenarios.

\end{abstract}

\begin{keywords}
    Federated learning, FedAVG, Scene classification, Remote sensing
\end{keywords}

\section{Introduction}
\label{sec:intro}
Remote sensing applications often deal with the challenge of processing vast amounts of data collected from, and thus distributed across geographical locations. Locally rounding up collected data for training a centralized model may infringe privacy regulations while training separate local models results in sub-optimal performance due to geographically non-uniform data distribution and overfitting. As the demand for sophisticated image classification models grows, the need for efficient and privacy-preserving methodologies becomes imperative.

Federated Learning (FL) emerges as a potential solution to enforce privacy by allowing decentralized devices to collaboratively train machine learning models across a distributed network while keeping raw data localized. Several approaches such as FedAVG, FedSGD, FedBN, etc.~\cite{li2021survey} and even a peer-to-peer (P2P)~\cite{toofanee2023federated} model have been proposed for FL, among which, FedAVG~\cite{mcmahan2017communication}, or Federated Averaging, is well-known for its efficiency and simplicity, acting as a baseline for other algorithms.
FedAVG has also been adopted for aggregating models across devices for remote sensing~\cite{buyuktacs2023learning} as
remote sensing tasks are usually run on devices located far from the ground, limiting the information being communicated \cite{lu2019compressed}, demanding that FL methods operating on small amount of information.

In this paper, we explore various FL communication strategies for remote sensing image classification based on the well-known FedAVG algorithm and study the trade-off between reducing data communicated and performances.
In particular, we propose sending and receiving only the average feature vectors for each category of interest, thus incurring minimal exchanged information.
As such, one client could gain insights from the features extracted by the other clients without explicitly synchronizing the trained weights among them.
The idea that a model can learn from feature representations instead of using network parameters has been widely applied between images and text as CLIP \cite{radford2021learning} employs Vision Transformer (ViT) to learn feature representations of text extracted from BERT, aiming to maximize the similarity of (image, text) pairs using a contrastive loss.
To that end, the Large Margin Cosine Loss \cite{wang2018cosface} is employed to solve the data imbalance problem and allow faster training convergence.
In regard to improving network performance, we combine the FedAVG algorithm with average feature communication and formulate the classification problem as a retrieval problem, obtaining the best of both worlds, high performance with fast convergence.

\section{Method}
\label{sec:method}

\subsection{Revisiting FedAVG}

This section briefly revisits and explains the FedAVG algorithm~\cite{mcmahan2017communication}. For more details, we refer interested readers to the original work.
FedAVG is a server-client communication model. A \emph{client} $C_i$ is an individual device training a local model using a separate set of data, disjoint with the others and the \emph{server} $S$ is the central entity responsible for maintaining a global model by aggregating the local models contributed from the clients. The training process composes of $t$ rounds of training the clients locally and communicating (for synchronization) between each of them and the server.

In FedAVG, each client, after $k$ epochs of local training, transmits all the trained parameters, also called \textit{weights} or, collectively, \textit{model}, to the server. The server's and all clients' models are, then, updated with the average model. The process is described in Algorithm~\ref{alg:algo1}.

\begin{algorithm}
\begin{small}
\caption{FedAVG training process}
\label{alg:algo1}
\begin{algorithmic}[1]
\LeftComment{Client's side}
\State Train model locally for $k$ epochs.
\State Send the updated model to the server $S$.
\LeftComment{Server's side}
\State Receive from all clients and compute the average model.
\State Send the average model to all clients.
\State Updated with the average model.
\LeftComment{Client's side}
\State Receive and updated with the average model.
\end{algorithmic}
\end{small}
\end{algorithm}

For evaluation, the server's global model is used and the prediction score from the softmax layer is used to determine the predicted category of an input image.

\subsection{Feature-based FedAVG}

To limit the amount of transferred information, we propose transferring only the backbone features of each client instead of the entire network's parameters which is costly. To that end, the Large Margin Cosine Loss (LMCL)~\cite{wang2018cosface} is employed in place of the Cross-Entropy loss with Softmax function commonly used in classification problems. We first describe the LMCL then explain how it facilitates the replacement of full model by extracted features.

\subsubsection*{Large Margin Cosine Loss}

The Cross-Entropy loss with Softmax function is described in Equation~\ref{eq:CEwithSoftmax},
where ${{y}_{i}}$ is the class to which the data ${{i}^{th}}$ belongs, $n$ is the number of labels, and $N$ is the number of samples.
\begin{equation}
L_{\text{Softmax}}=-\frac{1}{N}\sum\limits_{i=1}^{N}{\log }\frac{{{e}^{W_{{{y}_{i}}}^{T}{{x}_{i}}+{{b}_{{{y}_{i}}}}}}}{\sum\limits_{j=1}^{n}{{{e}^{W_{j}^{T}{{x}_{i}}+{{b}_{j}}}}}}.
\label{eq:CEwithSoftmax}
\end{equation}

The output logit of a network's backbone (at the \textit{head layer}, before Softmax), $W_j^Tx_i$, can be expressed using cosine function and the angle $\theta_{ji}$ between the vector $W_j^T$ and the feature $x_i$ as
\begin{equation}
W_{j}^{T}{{x}_{i}}=\left\| {{W}_{j}} \right\|\left\| {{x}_{i}} \right\|\cos \left( {{\theta }_{ji}} \right).
\end{equation}

Assuming that 
${{b}_{j}}=0$, $\left\| {{W}_{j}} \right\|=\left\| {{x}_{i}} \right\|=1$ (with $\left\|\cdot\right\|$ is the $l_{2}$ normalization \cite{wang2017normface}),
the original function (\ref{eq:CEwithSoftmax}) becomes:

\begin{equation}
\label{eq:LMCL1}
    L=-\frac{1}{N}\sum\limits_{i=1}^{N}{\log }\frac{{{e}^{\cos {{\theta }_{{{y}_{i}i}}}}}}{{{e}^{\cos {{\theta }_{{{y}_{i}i}}}}}+\sum\limits_{j=1,j\ne {{y}_{i}}}^{n}{{{e}^{\cos {{\theta }_{ji}}}}}}.
\end{equation}

Subsequently, the ArcFace algorithm \cite{deng2019arcface} introduces a margin value $m$ and a scalar parameter $s$ to increase performances, making LMCL function as follows:
\begin{equation}
\label{arcface}
    L_{ArcFace}=-\frac{1}{N}\sum\limits_{i=1}^{N}{\log }\frac{{{e}^{s\left( \cos \left( {{\theta }_{{{y}_{i}i}}}+m \right) \right)}}}{{{e}^{s\left( \cos \left( {{\theta }_{{{y}_{i}i}}}+m \right) \right)}}+\sum\limits_{j=1,j\ne {{y}_{i}}}^{n}{{{e}^{s\cos {{\theta }_{ji}}}}}}.
\end{equation}

\subsubsection*{Feature communication}
\label{method1}

LMCL is similar to the cross-entropy loss with softmax function but without requiring the head layer weights $W_j^T$. Instead, it replaces the weight by another feature vector and measures the similarity between them.
Therefore, is it possible to have clients exchange only the representations of features for each label with the server, without transmitting their weights. Algorithm~\ref{alg:algo2} describes the process.
\begin{algorithm}
\begin{small}
\caption{Communicating only average feature vectors}
\label{alg:algo2}
\begin{algorithmic}[1]
\LeftComment{Client's side}
\State Train model locally for $k$ epochs. \label{alg:algo2:begin-client}
\State Extract feature vectors before head layer for all training images\label{alg:algo2:extract-feat}
\State Compute average feature vectors for each category\label{alg:algo2:client-average}
\State Send the average vectors to the server.\label{alg:algo2:client-send-average}
\LeftComment{Server's side}
\State Receive and compute per-category average vectors of all clients.
\State Send the average vectors to each client. \label{alg:algo2:end-server}
\LeftComment{Client's side}
\State Receive and assign the average vectors as  weights for head layer.
\end{algorithmic}
\end{small}
\end{algorithm}

\begin{figure}
    \centering
    \includegraphics[]{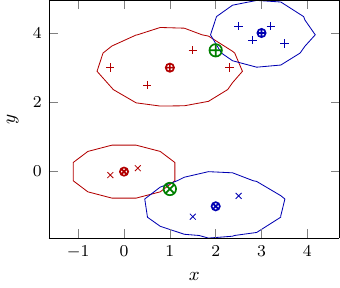}
    \caption{Visualization of Algorithm~\ref{alg:algo2} with 2 categories (indicated by $+$ and $\times$) and 2 clients (in red and blue). The per-class mean feature vectors of each category ($\oplus$ and $\otimes$) is computed by each client and sent to the server which computes the average mean vectors (in green) before
sending back to all the clients.}
    \label{fig:enter-label}
\end{figure}

Different from the vanilla FedAVG,
each client does not send the entire trained parameters to the server but only the averages of
the feature vectors extracted by the backbone with shape $\left[n_{\text{cat}}, d\right]$ (where $n_{\text{cat}}$ is the number of categories, and $d$ is the embedding dimension of the features).
The server receives $n_{\text{cat}}$ vectors from all the clients and compute $n_{\text{cat}}$ average features across the clients. Figure~\ref{fig:enter-label} shows a visualization.

In this way, only the average features for each label, which serve as representations of each class across all the clients, are transferred, maintaining minimal information communicated. The LMCL compels the clients' models to learn a representation that is close to these common representations. %

In this algorithm, the Server does not maintain an aggregated global model but only act as an intermediary to synchronize all the clients. For evaluation, we run all client models through the same validation data and average their performance.%

\subsubsection*{Regularization}
\addtocounter{algorithm}{1}
\dummylabel{alg:algo3}{\thealgorithm}

This method, however, may result in each client learning diverging representations from one another due to non-uniform variations of the local datasets and random weight initialization.
To constrain the deviation of the extracted features a regularization method is introduced.
In particular, the same set of random initial parameters denoted by $M_0$ is used to initialize all clients' backbones on the first round and modulate the backbones' learned parameters of each client as follows:

\begin{equation}
    \label{eq:reg}
    M_i = \dfrac{\left(n_{\text{client}} - 1\right) \times M_0 + \tilde{M}_{i}}{n_{\text{client}}},
\end{equation}

\noindent where $M_i$ is the backbone's set of parameters of a client after round $i$ and $\tilde{M}_i$ is the parameter set updated from the backpropagation. Equation~\ref{eq:reg} is used after Algorithm~\ref{alg:algo2}, before starting a new round. Subsequently, all the clients start from the same set of random parameters $M_0$ and the newly learned parameters are scaled down by $\frac{1}{n_{\text{client}}}$, which will be referred to as Algorithm~\thealgorithm.

\subsection{Model-and-features based FedAVG}
As it will be shown in the Section~\ref{sec:exp}, transferring only average features as in Algorithm~\ref{alg:algo2} results in fast converging, yet with the cost of inferior performance. Therefore, in this section, we propose to combine it to the original FedAVG in Algorithm~\ref{alg:algo1} and Algorithm~\ref{alg:algo2} to obtain the best of both worlds. The Algorithm~\ref{alg:algo1} is first run so that all clients are synchronized to the same parameters before the average feature vectors are sent and assigned to the clients' head layers as in Algorithm~\ref{alg:algo2}. The process is described in Algorithm~\ref{alg:algo4}.

\begin{algorithm}
\begin{small}
\caption{Communicating model and average features}
\label{alg:algo4}
\begin{algorithmic}[1]
\LeftComment{Training with FedAVG}
\State \textbf{Algorithm~\ref{alg:algo1}}
\LeftComment{Server and Clients: Synchronization features}
\State Perform steps 2 to 8 similar to \textbf{Algorithm~\ref{alg:algo2}}
\end{algorithmic}
\end{small}
\end{algorithm}

\subsection{Retrieval-based classification}
So far, classification prediction is provided by the softmax score of either the server's or clients' models and feature communication is only performed during the training process. In this section, we propose transferring the client's extracted features for making prediction during the deploying stage. To that end, the classification problem is reformulated as a retrieval problem and presented in Algorithm~\ref{alg:algo5}.
\begin{algorithm}
\begin{small}
\caption{Inference with extra features}
\label{alg:algo5}
\begin{algorithmic}[1]
\LeftComment{Training stage}
\State \textbf{Algorithm~\ref{alg:algo4}}
\LeftComment{Preparing for predicting stage}
\LeftComment{Client's side}
\State Extract backbone feature vectors for all \textit{training} images
\State The clients send \textit{all} feature vectors and labels to the server.\label{alg:algo5:client-send-all}
\LeftComment{Server's side}
\State Receive and transfer all feature vectors and labels to all clients.
\LeftComment{Client's side}
\State Receive and fit a kNN structure to all features and labels
\LeftComment{Predicting stage, client's side}
\State For each input image, extract \textit{backbone} feature vectors, and use kNN to obtain predicted category.
\end{algorithmic}
\end{small}
\end{algorithm}

The idea requires a preparation step before prediction can be made, which includes re-passing all training images through each client's backbone to extract the feature vectors, similar to step~\ref{alg:algo2:extract-feat} in Algorithm~\ref{alg:algo2}. The vectors are distributed to all other clients via the server. Each client classifies an input image by extracting its features using the backbone model and retrieving the majority label of the nearest training features in the embedding space.

\addtocounter{algorithm}{1}
\dummylabel{alg:algo6}{\thealgorithm}

Note the difference with Algorithm~\ref{alg:algo2}'s idea where the extracted feature of each training image is transferred instead of the per-class average vectors. As this could be expensive for large training sets, we also study the case where the average vectors (as in lines~\ref{alg:algo2:client-average},~\ref{alg:algo2:client-send-average} of Algorithm~\ref{alg:algo2}) are communicated and indicate it as Algorithm~\thealgorithm.
This will reduce the amount of information communicated across the network while being more secure \cite{zhu2023privacy} due to the low risk of data being exposed in case the features can be reversed.

\begin{figure}[ht]
    \centering
    \includegraphics[width=0.7\linewidth]{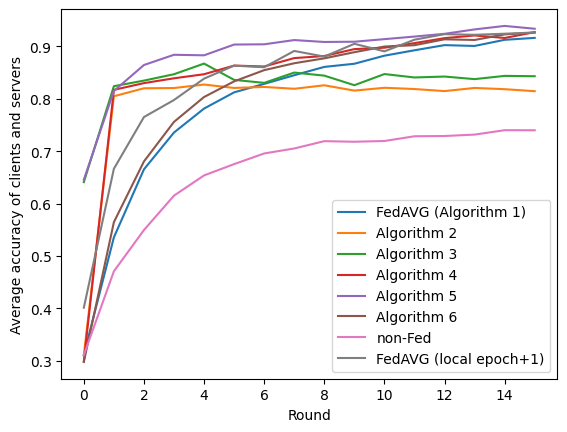}
    \caption{Comparative results when running Algorithms 1, 2, 3, 4, 5, and 6 on the UCM dataset. }
    \label{fig:ucm}
\end{figure}
\section{Experiments}
\label{sec:exp}

\subsection{Datasets and Setup}
In this work, we use two remote sensing scene classification datasets to conduct our experiments, namely the UC-Merced (UCM) \cite{yang2010bag} containing 100 images of size $256\times 256$ per class with 21 land-use classes and the Aerial Image Dataset (AID) \cite{xia2017aid} with $10,000$ images of size $600\times 600$ over $30$ scene classes. We use the UCM dataset for quick evaluation and AID for more analyses.
The Scikit-learn\footnote{[Online] Scikit-learn. URL \url{https://scikit-learn.org/}. Visisted on 25 May, 2024.}'s \texttt{train\_test\_split} function is used to split the dataset into training and validation set with \texttt{test\_size} of 0.3.

In all experiments, we use the same random seed, batch size of 16, learning rate of $10^{-4}$.
Server-client communication is run for $t=16$ rounds while client local training is run for 1 epoch with Adam optimization.
The number of clients is set to 10 for UCM and 20 for AID.
The ResNet18 model with ImageNet pre-trained weights is used as the network backbone with one fully connected layer added at the end to reduce the feature dimension to $128$.
Input images are all resized to $256 \times 256$ before feeding into the network. For ArcFace~\cite{deng2019arcface}, the margin $m=0.2$ and scale $s=20$ are chosen.

Regarding the non-identically distributed client data (non-IID), in the context of remote sensing, data distribution among clients is uneven due to each device being located at a specific geographical position. We draw samples from the Dirichlet distribution to split the dataset for each client.

\subsection{Results}
Figure \ref{fig:ucm} shows the results obtained for the UCM dataset with different FL strategies.
The non-Fed approach where the clients are trained independently without any communication is shown for reference purposes. The two feature-based Algorithms~\ref{alg:algo2} and~\ref{alg:algo3} converge more rapidly than the baseline FedAVG (blue) yet under-perform it in the long run. The superior of Algorithm~\ref{alg:algo3} over~\ref{alg:algo2} shows the benefit of regularization.
Combining feature and model-based communication benefits Algorithm~\ref{alg:algo4} and~\ref{alg:algo5}
in both ways, surpassing the baseline and maintaining till the final round. The retrieval method for obtaining predicted labels in Algorithm~\ref{alg:algo5} show further performance improvement.%

\begin{figure}[ht]
    \centering
    \includegraphics[width=0.7\linewidth]{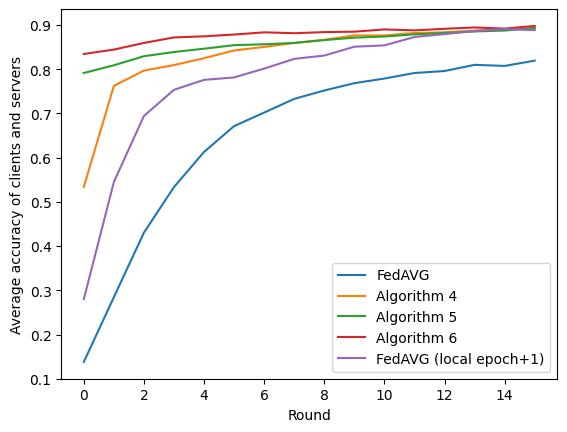}
    \caption{Comparative results when running Algorithms 4, 5, 6 on the AID dataset.}
    \label{fig:aid}
\end{figure}

The results obtained for the AID dataset are shown in Figure \ref{fig:aid}.
Since Algorithm~\ref{alg:algo5} and~\ref{alg:algo6} involve an extra passing of training data through clients' models to obtain anchor features for kNN retrieval although not requiring gradients and backpropagation, we include the baseline result with extra local epoch per round. Although being over-compensated, the baseline FedAVG with extra local epoch still under-performs the proposed Algorithm~\ref{alg:algo4},~\ref{alg:algo5},~\ref{alg:algo6}, and on par at the final round. %

The communication cost of each strategy is shown in Table \ref{table}. It can be seen that Algorithm~\ref{alg:algo2} and~\ref{alg:algo3} involve minimum data transmission and do not depend on model size (as there are no $w$ in their computation) while the other algorithms involve extra cost compared to the baseline.
\begin{table}[h]
\centering
\resizebox{\linewidth}{!}{\begin{tabular}{|c|c|c|c|}
\hline
{\bf Algorithm}   & {\bf Computation}                               & {\bf real bytes (UCM)} & {\bf real bytes (AID)} \\ \hline
1 & $A = 2\times w \times n_{clients}$                                 & 899,876,080                       & 1,799,752,160       \\ \hline
2 & $B = 2\times d \times n_{classes} \times n_{clients}$               & 215,040                          & 614,400           \\ \hline
3 & $B$                                                                & 215,040                          & 614,400           \\ \hline
4 & $A+B$                                                              & 900,091,120                       & 1,800,366,560       \\ \hline
5 & $A+n_{samples} \times \left(2+ d \times n_{clients}^2\right)$ & 975,156,880                       & 3,233,432,160       \\ \hline
6 & $A+ n_{clients} \times B$                                                   & 902,026,480                       & 1,812,040,160       \\ \hline
\end{tabular}}
\caption{Total data sent in 1 network training round (in bytes). $w$ denotes Resnet18 weights. $d$ denotes the backbone feature dimension. $n_{clients}$, $n_{classes}$ and $n_{samples}$ denote the number of clients, classes and samples, respectively.}
\label{table}
\end{table}

\section{Conclusion}
\label{sec:conclusion}
This work has proposed and compared different FL approaches, including communicating only with features, combining features with pseudo-weights, and utilizing both weights and features for training and testing. Our experiments on remote sensing two datasets showed the effectiveness of these approaches by highlighting faster convergence and reducing the amount of information transmitted over the network. This research contributes to exploring the impacts of features in various communication strategies in FL, particularly in the context of remote sensing applications.

 \newpage
{\small
\bibliographystyle{ieeetr}
\bibliography{macro,IRISA-IGARSS23}
}

\end{document}